%% file: ICASSP2021.tex
\documentclass{article}
\usepackage{spconf,amsmath,graphicx}
\usepackage{times}
\usepackage{url}
\usepackage{latexsym}
\usepackage{amsmath,amssymb,amsfonts}
\usepackage{algorithmic}
\usepackage{graphicx}
\usepackage{textcomp}
\usepackage{comment}
\usepackage{xcolor}
\usepackage{CJK}
\usepackage{color}
\usepackage{booktabs}
\usepackage{multirow}
\usepackage{multicol}
\usepackage{mathrsfs}

\title{Learning to Select Context in a Hierarchical and Global Perspective for Open-domain Dialogue Generation}
\name{Lei Shen$^{1,2,*}$, Haolan Zhan$^{2,*}$, Xin Shen$^3$, Yang Feng$^{1,2}$\thanks{$1$ IIP stands for Key Laboratory of Intelligent Information Processing.}\thanks{$*$ Equal Contribution. Corresponding to: shenlei17z@ict.ac.cn}}
\address{$^1$IIP, Institute of Computing Technology, Chinese Academy of Sciences \\ $^2$University of Chinese Academy of Sciences, $^3$Australian National University}
\begin{document}
%\ninept
%
\maketitle
\input{abstract}

\begin{keywords}
Open-domain Dialogue Generation, Context Selection, Hierarchical and Global Perspective
\end{keywords}
\input{introduction}

\input{model}

\input{experiments}
\input{conclusion}

\vfill\pagebreak

\bibliographystyle{IEEEbib}
%\bibliography{strings,refs}
\bibliography{coling2020}

% \newpage

% \input{rebuttal/review 2877}

% \newpage

% \input{rebuttal/review 07AA}

% \newpage
% \input{rebuttal/review 0099}

\end{document}

%% file: abstract.tex
\begin{abstract}
Open-domain multi-turn conversations mainly have three features, which are hierarchical semantic structure, redundant information, and long-term dependency. Grounded on these, selecting relevant context becomes a challenge step for multi-turn dialogue generation. However, existing methods cannot differentiate both useful words and utterances in long distances from a response. Besides, previous work just performs context selection based on a state in the decoder, which lacks a global guidance and could lead some focuses on irrelevant or unnecessary information. In this paper, we propose a novel model with hierarchical self-attention mechanism and distant supervision to not only detect relevant words and utterances in short and long distances, but also discern related information globally when decoding. Experimental results on two public datasets of both automatic and human evaluations show that our model significantly outperforms other baselines in terms of fluency, coherence, and informativeness.
\end{abstract}

%% file: introduction.tex
\section{Introduction}
% introduction缩减到第一页内
Open-domain multi-turn dialogue generation has gained increasing attentions in recent years, as it is more accordant with real scenarios and aims to produce customized responses. In general, an open-domain multi-turn conversation has following features: (1) The context (including the query and previous utterances in our paper) is in a hierarchical structure, which means it consists of some utterances, and each utterance contains several words. (2) At most cases, many contents of the context are redundant and irrelevant to the response. (3) Some related information (utterances or words) and the response are in a long-term dependency relation. Therefore, \textit{Context Selection}, detecting the relevant context based on which to generate a more coherent and informative response, is a key point in multi-turn dialogue generation.

Based on feature (1), the hierarchical recurrent encoder-decoder network (HRED) \cite{serban2016building} has been proposed. It encodes each utterance and the whole context at two levels, and is widely applied to other methods for multi-turn dialogue generation. Then, hierarchical recurrent attention \cite{song2020hka} and explicit weighting \cite{tian2017make,zhang2018context}, memory networks \cite{chen2018hierarchical} and self-attention mechanism \cite{zhang2019recosa} have been introduced to match feature (2) and (3), respectively. However, few work could cover all these features simultaneously to fulfill context selection and response generation tasks. 

When it comes to \textit{Context Selection}, existing methods can be categorised into two ways: (1) Detecting related utterances measured by the similarity between query and each previous utterance \cite{tian2017make,zhang2018context}. (2) Applying the attention mechanism from a local perspective, i.e., based solely on the current state in decoder with the Maximum Likelihood Estimation (MLE) loss \cite{zhang2018context,zhang2019recosa}. The similarity measurement in the former cannot select word-level context, while the guidance from the local perspective in the latter would make the model choose some deviated context and produce an inappropriate response \cite{shen2019modeling,shen2020cdl,shen2020compose}.
% \cite{ren2019thinking}. 

To tackle the above mentioned problems, we propose \textbf{HiSA-GDS}, a modified Transformer model with \textbf{Hi}erarchical \textbf{S}elf-\textbf{A}ttention and \textbf{G}lobally \textbf{D}istant \textbf{S}upervision. To the best of our knowledge, it is the first time to design these two modules for open-domain dialogue generation. Specifically, we use Transformer encoder to encode each utterance in the context. During training, the response is firstly processed by a masked self-attention layer, and then a word-word attention aggregates related word information in each utterance individually. After that, we conduct utterance-level self-attention to get context-sensitive representations of aggregated information from last layer. Then, we calculate the attention weights between utterance-level outputs of the previous layer and the masked response representation. Finally, we generate the corresponding response based on the fusion of selected information at both word and utterance levels. Besides, to provide a global guidance of decoding, we import a distant supervision module which utilizes the similarity score between the response and each contextual utterance measured by a pre-trained sentence-embedding model. All parameters are learned based on the global Distant Supervision and local MLE in an end-to-end framework. 

Experimental results on two public datasets along with further discussions show that HiSA-GDS significantly outperforms other baselines and is capable to generate more fluent, coherent, and informative responses.

%% file: model.tex
\section{Approach}
\begin{figure}
\centerline{\includegraphics[width=1.0\linewidth]{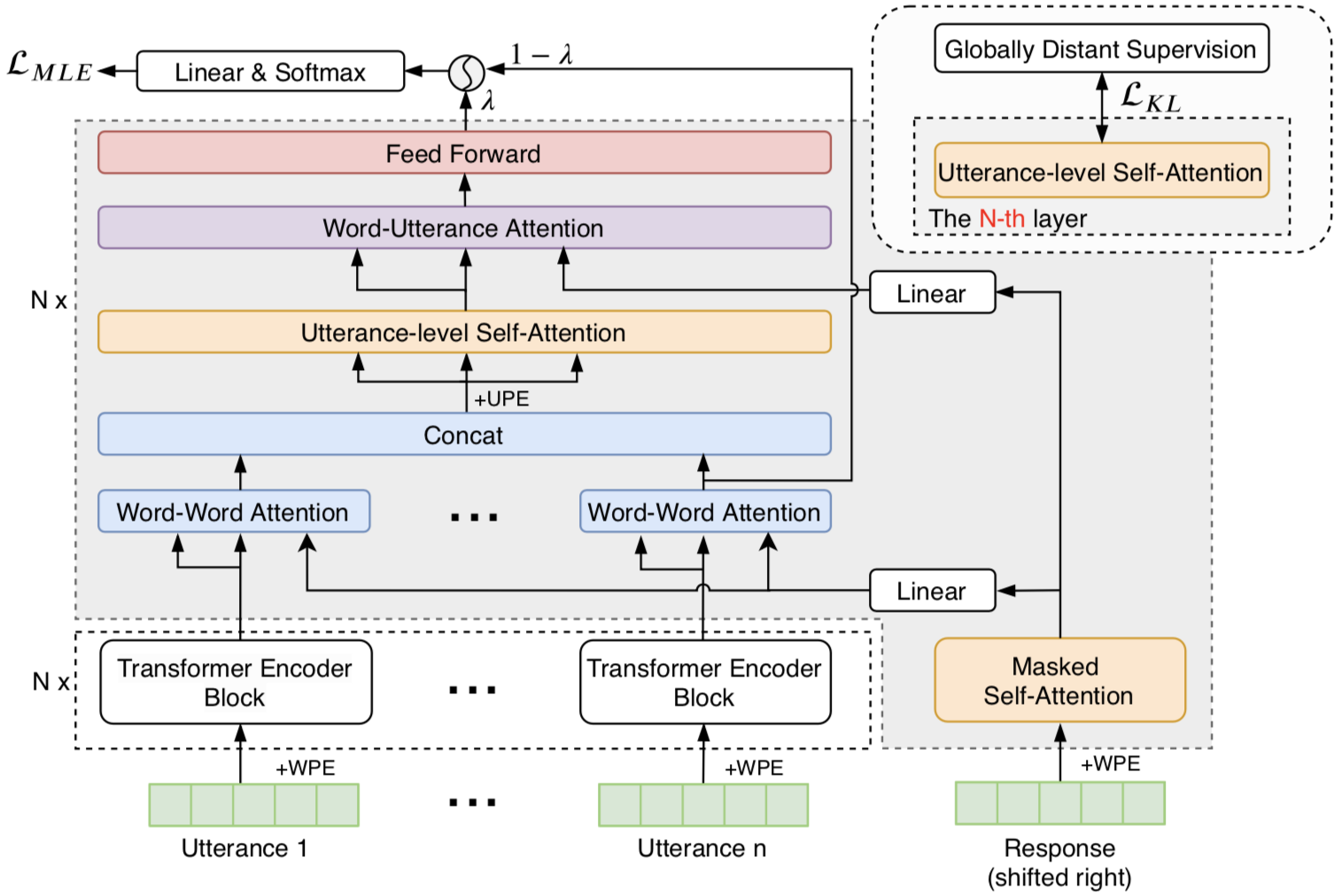}}
\vspace{-3mm}
\caption{Architecture of HiSA-GDS. The white dashed box is Transformer encoder, while the gray one is the modified Transformer decoder. The residual connection and layer normalization are omitted for brevity. ``WPE'' and ``UPE'' represent word position encoding and utterance position encoding. The upper right corner shows the globally distant supervision that is only introduced to the $N$-th layer of decoder.}
\label{fig:model_fig}
\end{figure}

The input is a context containing $n$ utterances $\{{X}_i\}_{i=1}^{n}$, and each utterance is defined as $X_i = \{x_{i,1},...,x_{i,|X_i|}\}$, where $|X_i|$ is the length of the $i$-th utterance and $x_{i,m}$ is the $m$-th word of $X_i$. Our goal is to select relevant context consisting of utterances and words, and then generate a response $Y$ = \{$y_1, y_2, ..., y_{|Y|}$\} by utilizing the related information, where $|Y|$ is the length of response $Y$.

\subsection{Encoder}
We consider each utterance independently, and given an utterance $X_i$, the input representation of word $x_{i,j}$ is the sum of its word embedding and position encoding: $I(x_{i,j}) = {\rm WE}(x_{i,j}) + {\rm WPE}(x_{i,j})$,
% \begin{equation}
%     I(x_{i,j}) = {\rm WE}(x_{i,j}) + {\rm WPE}(x_{i,j}),
% \end{equation}
where ${\rm WE}(x_{i,j})$ and ${\rm WPE}(x_{i,j})$ represent word and word position embedding, respectively. The input embedding is then fed into Transformer encoder with $N$ layers. 
% Each layer contains two sub-layers, which is a self-attention layer followed by a feed-forward neural network (FFN).
% :
% \begin{align}
%   & \mathbf{E}^{(0)}_i = [I(x_{i,1}), I(x_{i,2}), ... , I(x_{i,|X_i|})],\\
%   & \mathbf{E}^{(l)}_i = {\rm FFN}({\rm MHA}(\mathbf{E}^{(l-1)}_i, \mathbf{E}^{(l-1)}_i, \mathbf{E}^{(l-1)}_i)),
% \end{align}
% where [$\cdot$] denotes the concatenation operation, and $\mathbf{E}^{(0)}_i$ is the initial input representation at the first layer. The sub-layer ${\rm FFN}$ is a position-wise fully connected feed-forward network and ${\rm MHA}(\mathbf{Q},\mathbf{K},\mathbf{V})$ is a multi-head attention function taking a query matrix $\mathbf{Q}$, a key matrix $\mathbf{K}$, and a value matrix $\mathbf{V}$ as inputs. 
The final encoding of $X_i$ is the output from the $N$-th layer, $\mathbf{E}^{(N)}_i$. Please refer to \cite{vaswani2017attention} for more details.

\subsection{Hierarchical Self-Attention based Decoder}
The decoder also contains $N$ layers, and each layer is composed of five sub-layers. The first sub-layer is a masked self-attention, which is defined as:
\begin{equation}
    \mathbf{M}^{(l)}_t = {\rm MHA}(\mathbf{D}^{(l-1)}_t, \mathbf{D}^{(l-1)}_t, \mathbf{D}^{(l-1)}_t),
\end{equation}
where ${\rm MHA}$ is the multi-head attention function, $\mathbf{D}^{(l-1)}_t$ denotes the input representation of the $l$-th layer, and $\mathbf{M}^{(l)}_t$ denotes the output of masked self-attention at the $l$-th layer. $\mathbf{D}^{(0)}_t$ is the concatenated result of all words before time step $t$ in the response and each word is also represented as the sum of its word embedding and position encoding. 

The second sub-layer is a word-word attention that summarizes word-level response-related information from each utterance $X_i$ into a vector at a specific decoding time:
\begin{equation}
    \mathbf{U}^{(l)}_{t,i} = {\rm MHA}(f_w(\mathbf{M}^{(l)}_t), \mathbf{E}^{(N)}_i, \mathbf{E}^{(N)}_i),
\end{equation}
where $f_w$ is a linear transformation.

The third sub-layer is an utterance-level self-attention. Inspired by Zhang et al. \cite{zhang2019recosa}, we also utilize the self-attention mechanism to capture the long-term dependency of utterance-level information. Similar to word position encoding, we add utterance position encoding ($\rm UPE$) to $\mathbf{U}^{(l)}_{t,i}$, and denote the sum result as $\mathbf{\Tilde{U}}^{(l)}_{t,i}$. The output of this sub-layer is calculated as:
\begin{equation}
    \mathbf{H}^{(l)}_t = {\rm MHA}(\mathbf{\Tilde{U}}^{(l)}_t, \mathbf{\Tilde{U}}^{(l)}_t, \mathbf{\Tilde{U}}^{(l)}_t),
\end{equation}
where $\mathbf{\Tilde{U}}^{(l)}_t = [\mathbf{\Tilde{U}}^{(l)}_{t,1}, \mathbf{\Tilde{U}}^{(l)}_{t,2},...,\mathbf{\Tilde{U}}^{(l)}_{t,n}]$. Then, the fourth sub-layer is a word-utterance attention layer to find out utterance-level relevant information which is defined as:
\begin{equation}
    \mathbf{C}^{(l)}_t = f_l({\rm MHA}(f_u(\mathbf{M}^{(l)}_t), \mathbf{H}^{(l)}_t, \mathbf{H}^{(l)}_t)),
\end{equation}
where $f_l$ and $f_u$ are linear transformations, and $f_l$ is used for changing the output dimension. The last sub-layer is a feed-forward neural network (FFN):
\begin{equation}
    \mathbf{F}^{(l)}_t = {\rm FFN}(\mathbf{C}^{(l)}_t).
\end{equation}

Each of above mentioned sub-layer is followed by a normalization layer and a residual connection. Finally, we use a fusion gate to regulate the relevant information at word level ($\mathbf{U}^{(l)}_{t,n}$) and utterance level ($\mathbf{F}^{(l)}_t$): 
\begin{align}
    \lambda_t &= \sigma(W_g[\mathbf{U}^{(l)}_{t,n}, \mathbf{F}^{(l)}_t]), \\
    \mathbf{D}^{(l)}_t &= \lambda_t \ast \mathbf{F}^{(l)}_t + (1-\lambda_t) \ast \mathbf{U}^{(l)}_{t,n},
\end{align}
where $W_g$ is parameter metric, $\sigma$ is the sigmoid activation function, and $\ast$ means the point-wise product.

\subsection{Globally Distant Supervision}

Previous attention-based models achieve context selection from a local perspective, i.e., they try to generate one token at a time based solely on the current decoding state, which would detect deviated context and mislead the further generation. Besides, we do not have manual annotations to provide direct signals for selection. To address these problems, we design a globally distant supervision module to help determine relevant information, which provides a global guidance for the response generation process. Firstly, we apply a high quality pre-trained sentence-embedding model to encode contextual utterance $X_i$ and response $Y$ into vectors, denoted as $\mathbf{x}_i$ and $\mathbf{y}$. Then, we use the dot product to measure the semantic relevance between $\mathbf{x}_i$ and $\mathbf{y}$ \cite{lian2019learning}, and compute the selection probability as follows:
\begin{equation}
    \label{qequation}
    {P}\left(\mathbf{x} = \mathbf{x}_i|\mathbf{y}\right) = \frac{{\rm exp}(\mathbf{x}_{i}\cdot\mathbf{y})}{\sum_{j=1}^{n}{{\rm exp}\left(\mathbf{x}_j
    \cdot\mathbf{y}\right)}}.
\end{equation}

\subsection{Training Objective}

We utilize three loss functions in our training process. The first one is MLE loss which is defined as:
\begin{equation}
   \mathcal{L}_{MLE}(\theta) = -\frac{1}{|Y|}\sum_{t=1}^{|Y|} {\rm log} p(y_t|y_{<t}, \{{X}_i\}_{i=1}^{n};\theta),
\end{equation}
where $\theta$ represents the model parameters, and $y_{<t}$ denotes the previously generated words. Since MLE loss only provides local (token-wise) supervision, inspired by Ren et al. \cite{ren2019thinking} and Zhan et al. \cite{zhan2020user}, we apply the Kullback-Leibler divergence (KL) loss and the Maximum Causal Entropy (MCE) loss for globally distant supervision. KL loss measures the distance between two distributions: $P(\mathbf{x|y)}$, which is the distant ground-truth supervision described in Equation \ref{qequation}, and $Q(\mathbf{x|y)}=\frac{1}{|Y|}\sum_{t=1}^{|Y|}\mathbf{C}^{(N)}_t$, which is the average sum of estimated probabilities at all steps from the output of word-utterance attention sub-layer in the last decoder layer. We denote the KL loss as:
\begin{equation}
   \mathcal{L}_{KL}(\theta) = {\rm KL}(P(\mathbf{x}|\mathbf{y}) || Q(\mathbf{x}|\mathbf{y});\theta).
\end{equation}

Then, we use MCE loss to alleviate the negative effects of noises caused by imprecise $Q(\mathbf{x}|\mathbf{y})$:
\begin{equation}
    \mathcal{L}_{MCE}(\theta) = \frac{1}{|Y|}\sum_{t=1}^{|Y|}\sum_{w \in V}P(y_t=w){\rm log}P(y_t=w),
\end{equation}
where $V$ denotes the vocabulary. Finally, our overall loss is a linear combination of these three loss functions:
\begin{equation}
   \mathcal{L}(\theta) =
   \mathcal{L}_{MLE}(\theta) +
   \eta_1\mathcal{L}_{KL}(\theta) + \eta_2\mathcal{L}_{MCE}(\theta),
\label{e15}
\end{equation}
where hyper-parameters $\eta_1$ and $\eta_2$ govern the relative importance of different loss terms.

%% file: experiments.tex
\section{Experiment Settings}

\noindent \textbf{Datasets}:
We evaluate the performance on two public datasets: Ubuntu Dialogue Corpus \cite{lowe2015ubuntu} (\textit{Ubuntu}) and JD Customer Service Corpus \cite{chen2020jddc} (\textit{JDDC}).

\noindent \textbf{Baselines}: (1) Seq2Seq with Attention Mechanism ({\bf S2SA}) \cite{sutskever2014sequence}, and we concatenate all context utterances as a long sequence; (2) Hierarchical Recurrent Encoder-Decoder ({\bf HRED}) \cite{serban2016building}; (3) Variational HRED ({\bf VHRED}) \cite{serban2017hierarchical} with word drop and KL annealing, and the word drop ratio equals to 0.25; (4) Static Attention based Decoding Network ({\bf Static}) \cite{zhang2018context}; (5) Hierarchical Recurrent Attention Network ({\bf HRAN}) \cite{xing2018hierarchical}; (6) {\bf Transformer} \cite{vaswani2017attention}, and we concatenate all context utterances into a long sequence; (7) Relevant Contexts Detection with Self-Attention Model ({\bf ReCoSa}) \cite{zhang2019recosa}. They all focus on multi-turn conversations, and ReCoSa is a state-of-the-art model on both \textit{Ubuntu} and \textit{JDDC}. For ablation study, {\bf HiSA} is our model without the globally distant supervision.

\noindent \textbf{Hyper-parameters}: The utterance padding length is set to 30, and the maximum conversation length is 10. The hidden size of encoder and decoder is 512, and the number of layers is 4 for encoder and 2 for decoder. The head number of multi-head attention is set to 8. The high-quality pre-trained sentence-embedding model we used is Infersent \cite{conneau2017supervised}/Familia \cite{jiang2018familia} for {\it Ubuntu}/{\it JDDC}. These models are both pre-trained on large-scale datasets in either English or Chinese, and perform well on our datasets. For optimization, we use Adam \cite{kingma2014adam} with a learning rate of 0.0001 with gradient clipping. Hyper-parameters in Equation \ref{e15} are set to 1.

\noindent \textbf{Performance Measures}:
For automatic evaluation, we use 4 groups of metrics: (1) \textbf{BLEU-2} \cite{papineni2002bleu}; (2) {\bf Embedding-based Metrics} (Average, Greedy, and Extrema) \cite{serban2017hierarchical}; (3) {\bf Coherence} \cite{xu2018better} that evaluates the semantic coherence between the context and response; (4) {\bf Distinct-2} \cite{li2016diversity}. For human evaluation, we utilize the side-by-side human comparison. We invite 7 postgraduate students as annotators. To each annotator, we show a context with two generated responses, one from HiSA-GDS and the other from a baseline model, but the annotators do not know the order. Then we ask annotators to judge which one wins based on fluency, coherence, and informativeness. Please refer to \cite{xing2018hierarchical} for more details. Agreements among the annotators are calculated using Fleiss' kappa.
% ~\cite{fleiss1971}.

\section{Results and Discussion}
% \noindent \textbf{Results and Discussion}:

\begin{table*}[!htb]
\footnotesize
\centering
\begin{tabular}{c|cccccc|cccccc}
\toprule[1pt]
   \multirow{2}{*}{Model} & \multicolumn{6}{c|}{Ubuntu} & \multicolumn{6}{c}{JDDC} \\ \cline{2-13}
   ~ & B-2 & D-2 & Avg & Ext & Gre & Coh & B-2 & D-2 & Avg & Ext & Gre & Coh \\
  \hline
   S2SA \cite{sutskever2014sequence} & 0.896 & 6.104 & 46.323 & 28.851 & 39.209 & 48.117 & 4.233  & 3.609 & 53.901 & 36.493  & 37.578 & 46.176\\
   HRED \cite{serban2016building} & 3.853 & 6.661 & 57.972 & 34.007 & 41.462  & 63.173 & 9.405 & 11.762 & 63.191 & 46.714 & 43.295 & 57.183\\
   VHRED \cite{serban2017hierarchical} & 3.677 & 8.098  & 57.251 & 32.024  & 41.808 & 61.464 & 6.367 & 15.184 & 62.436 & 43.337 & 41.787 & 63.924\\
   Static \cite{zhang2018context} & 1.581 & 3.586 & 51.055 & 36.193 & \textbf{53.983} & 69.748 & 2.285 & 3.738 & 60.820 & 38.047 & 35.367 & 65.938 \\
   HRAN \cite{xing2018hierarchical} & 3.880 & 7.402 & 56.763 & 33.501  &41.584& 67.635  & 5.962 & \textbf{16.365} & 63.064 & 43.439 & 42.389 & 62.391\\
   Transformer \cite{vaswani2017attention}&  3.697 & 7.278 & 53.463 & 36.353 & 42.763 & 69.970 & 5.389 & 5.185 & 68.336 & 48.284 & 41.103 & 67.485\\
   ReCoSa \cite{zhang2019recosa} & 3.872 & 9.406 & 59.368 & 35.834 & 41.835 & 71.922 & 5.962 & 6.594 & 61.085 & 41.473 & 42.942 & 71.374\\ \hline
   HiSA & 4.021 & 9.598 & 63.527 & 36.208 & 40.598 & 72.261  &  6.986 & 14.804 & 66.103 & 43.715 & 45.081 & 73.286\\ 
  \textbf{HiSA-GDS} & \textbf{7.351} & \textbf{10.934} & \textbf{68.283} & \textbf{41.468} &  {50.382} & \textbf{75.823} & \textbf{7.127} & {15.823} & \textbf{73.952} & \textbf{52.502} & \textbf{49.477} & \textbf{74.281} \\
 \bottomrule[1pt]
\end{tabular}
\vspace{-2mm}
\caption{Automatic evaluation results on {\it Ubuntu} and {\it JDDC} (\%). The metrics BLEU-2, Distinct-2, Average, Extrema, Greedy and Coherence are
abbreviated as B-2, D-2, Avg, Ext, Gre, and Coh, respectively.}
\label{tab_auto-eval}
\end{table*}

\noindent\textbf{Automatic Evaluation Results:} As shown in Table \ref{tab_auto-eval}, our model outperforms all baselines significantly on both {\it Ubuntu} and {\it JDDC} (significance tests, $p$-value $<$ 0.01) by achieving the highest scores in almost all automatic metrics. Compared with existing baseline models, our model demonstrates its ability of generating relevant and appropriate responses. This is supported by the fact that results of our proposed model have gained improvements on BLEU-2, Embedding-based Metrics, and Coherence. Besides, we also achieve higher Distinct-2 score, which indicates that HiSA-GDS can generate more informative responses.

 \begin{table}[!t]
\centering
\footnotesize
\begin{tabular}{c|c|ccc|c}
% |c|c|ccc|c}
  \toprule[1pt] 
  \multirow{2}{0.9cm}{Dataset} & \multirow{2}{0.8cm}{Model} & \multicolumn{3}{c|}{HiSA-GDS vs.} & \multirow{2}{0.8cm}{kappa}\\ 
\cline{3-5} %\cline{9-11} 
  ~ & ~ & Win & Loss & Tie & ~ \\
%   \~ & ~ & Win & Loss & Tie & ~\\
  \hline
  \multirow{7}{*}{Ubuntu} & S2SA \cite{sutskever2014sequence} & 58\% & 12\% & 30\% & 0.468 \\
  ~ & HRED \cite{serban2016building} & 46\% & 19\% & 35\% & 0.531 \\
  ~ & VHRED \cite{serban2017hierarchical} & 48\% & 20\% & 32\% & 0.493 \\
  ~ & Static \cite{zhang2018context} & 51\% & 17\% & 32\% & 0.596  \\
  ~ & HRAN \cite{xing2018hierarchical} & 42\% & 9\% & 49\% & 0.424 \\
  ~ & Transformer \cite{vaswani2017attention} & 44\% & 19\% & 37\% & 0.474  \\
  ~ & ReCoSa \cite{zhang2019recosa} & 40\% & 6\% & 54\% & 0.528 \\ \hline
  \multirow{7}{*}{JDDC} & S2SA \cite{sutskever2014sequence} & 53\% & 24\% & 23\% & 0.547  \\
  ~ & HRED \cite{serban2016building} & 56\% & 16\% & 34\% & 0.468 \\
  ~ & VHRED \cite{serban2017hierarchical} &  52\% & 19\% & 29\% & 0.453  \\
  ~ & Static \cite{zhang2018context} & 48\% & 11\% & 41\% & 0.518 \\
  ~ & HRAN \cite{xing2018hierarchical} & 50\% & 22\% & 28\% & 0.495 \\
  ~ & Transformer \cite{vaswani2017attention} & 51\% & 29\% & 20\% & 0.447   \\
  ~ & ReCoSa \cite{zhang2019recosa} & 45\% & 27\% & 28\% & 0.461  \\
  \bottomrule[1pt]
\end{tabular}
\vspace{-2mm}
\caption{Human evaluation between HiSA-GDS and other baselines on \textit{Ubuntu} and \textit{JDDC}.}
\label{tab:human}
\end{table}

% \caption{Human evaluation between HiSA-GDS and other baselines on \textit{Ubuntu} and \textit{JDDC} datasets.}

\noindent\textbf{Human Evaluation Results:} These results are shown in Table \ref{tab:human}. We observe that HiSA-GDS outperforms all baseline models on both {\it Ubuntu} and {\it JDDC}. Specifically, the percentage of ``win'' is always larger than that of ``loss''. Take {\it Ubuntu} dataset as an example. Compared with VHRED and Transformer, HiSA-GDS achieves preference gains with 48\%, 51\%, and 44\%, respectively. We check responses generated by our model with ``win'' and find that they are more relevant to contextual utterances. The kappa scores indicate that annotators come to a ``Moderate agreement'' on judgement.

\begin{figure}
\centerline{\includegraphics[width=1.0\linewidth]{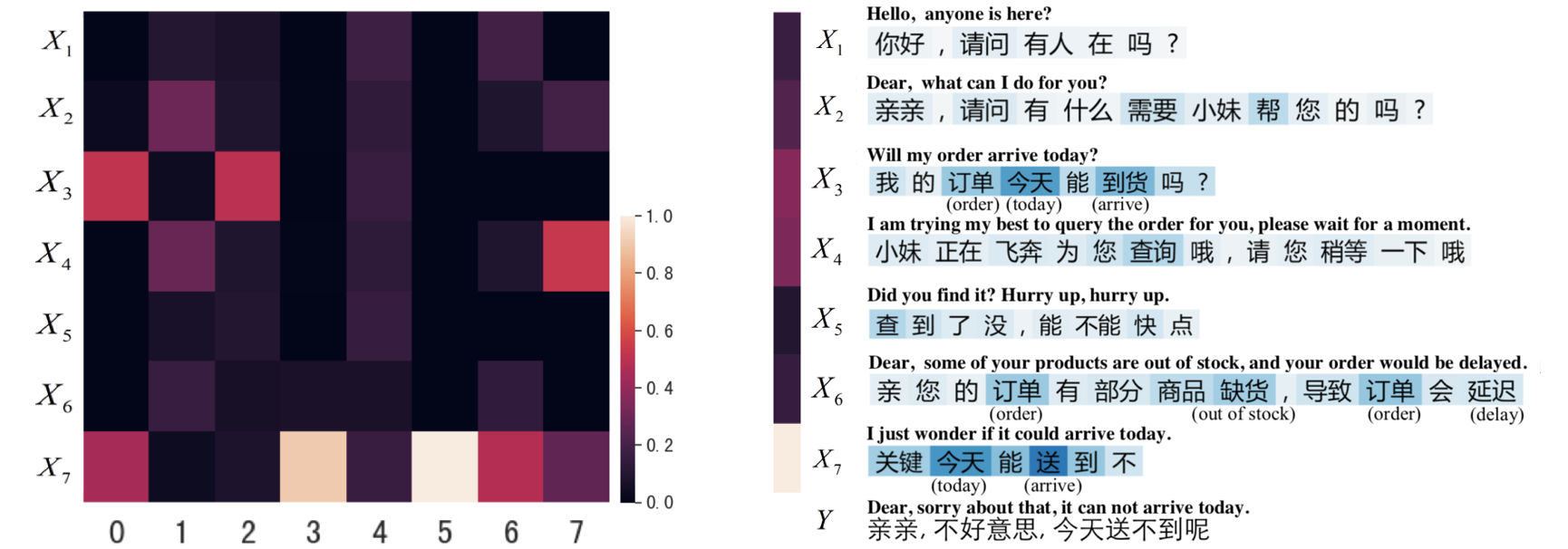}}
\vspace{-2mm}
\caption{Left: Utterance-level multi-head attention visualization of HiSA-GDS in the word-utterance attention layer. 0 to 7 are the index of each head. Right: Word-level attention visualization in the word-word attention layer. The importance of a word (horizontal blue heatmap) or an utterance (vertical red heatmap) is calculated as the average value of all heads.}
\label{fig:combine}
\end{figure}

\noindent\textbf{Discussion of Hierarchical Self-Attention:} To validate the effectiveness of hierarchical self-attention mechanism, we present the heatmap of an example in Figure \ref{fig:combine}. In this example, there are seven contextual utterances, and for each utterance, importance of each word is indicated with the depth of blue color on the right part. Besides, we also show an utterance-level attention visualization on the left part. An utterance is more important when the red color is lighter. For example, the third and seventh utterances, i.e., $X_3$ and $X_7$, are more important than the others. The importance of a word (horizontal heatmap on the right of $X_1$ to $X_7$) or an utterance (vertical heatmap on the left of $X_1$ to $X_7$) is calculated as the average value of different heads. From the word-level visualization, we find that words including ``\begin{CJK*}{UTF8}{gbsn}订单\end{CJK*}(order)'', ``\begin{CJK*}{UTF8}{gbsn}今天\end{CJK*}(today)'', and ``\begin{CJK*}{UTF8}{gbsn}送
货\end{CJK*}(deliver)'' are selected to be more relevant. Overall, the results are in accordance with humans' judgement and have achieved the goal of our proposed model. 

\noindent\textbf{Discussion of GDS:} Since GDS is only utilized during the training process, we calculate the relevance score between each contextual utterance and the ground-truth response. After applying Familia \cite{jiang2018familia} over the entire conversation, the relevance scores are 0.1502, 0.1388, 0.1602, 0.1548, 0.0979, 0.1343, and 0.1638 for $X_1$ to $X_7$, which is consistent with humans' intuition. Besides, inspired by Zhang et al. \cite{zhang2019recosa}, we randomly sample 300 context-response pairs from {\it JDDC}. Three annotators who are postgraduate students are invited to label each context. If a contextual utterance is related to the response, then it is labeled as 1. The kappa value is 0.568, which indicates the moderate consistency among different annotators. We then pick out samples that is labeled the same by at least two annotators, and then calculate the kappa value between humans' judgement and the outputs from Familia \cite{jiang2018familia} on these cases. The value 0.863 reflects ``Substantial agreement'' between them.

%% file: conclusion.tex
\section{Conclusion}
In this paper, we propose a novel model for open-domain dialogue generation, HiSA-GDS, which conducts context selection in a hierarchical and global perspective. The hierarchical self-attention is introduced to capture relevant context at both word and utterance levels. We also design a globally distant supervision module to guide the response generation at decoding. Experiments show that HiSA-GDS can generate more fluent, coherent, and informative responses.